\documentclass[journal]{IEEEtran}
\usepackage{amsmath,amssymb,amsfonts}
\usepackage{algorithmic}
\usepackage{xcolor}
\usepackage{graphicx}
\usepackage{textcomp}
\usepackage{booktabs} 
\usepackage{subfig}
\newcommand{\etal}{\textit{et al}. }
\usepackage{mathtools}
\usepackage{multirow}
\usepackage{graphics}
\usepackage{array}
\usepackage{comment}
\usepackage{threeparttable}
\usepackage{cite,etoolbox}

\def\BibTeX{{\rm B\kern-.05em{\sc i\kern-.025em b}\kern-.08em
    T\kern-.1667em\lower.7ex\hbox{E}\kern-.125emX}}
\begin{document}
\title{NSSI-Net: A Multi-Concept GAN for Non-Suicidal Self-Injury Detection Using High-Dimensional EEG in a Semi-Supervised Framework}
\author{Zhen Liang\textsuperscript{1,*}, \IEEEmembership{Member, IEEE}, Weishan Ye\textsuperscript{1}, Qile Liu, Li Zhang, Gan Huang, and Yongjie Zhou\textsuperscript{*}
\thanks{This work was supported by the National Natural Science Foundation of China (62276169), Medical-Engineering Interdisciplinary Research Foundation of Shenzhen University (2023YG004), Shenzhen University-Lingnan University Joint Research Programme, Shenzhen-Hong Kong Institute of Brain Science-Shenzhen Fundamental Research Institutions (2023SHIBS0003), the STI 2030-Major Projects (2021ZD0200500), the Open Research Fund of the State Key Laboratory of Brain-Machine Intelligence, Zhejiang University (Grant No. BMI2400008), and the Shenzhen Science and Technology Program (No. JCYJ20241202124222027 and JCYJ20241202124209011).}
\thanks{Zhen Liang and Weishan Ye are co-first authors. Zhen Liang, Weishan Ye, Qile Liu, Li Zhang, and Gan Huang are with the School of Biomedical Engineering, Medical School, Shenzhen University, Shenzhen 518060, China, and also with the Guangdong Provincial Key Laboratory of Biomedical Measurements and Ultrasound Imaging, Shenzhen 518060, China. E-mail: janezliang@szu.edu.cn, \{2110246024, liuqile2022\}@email.szu.edu.cn, and \{lzhang, huanggan\}@szu.edu.cn.}
\thanks{Yongjie Zhou is with the Department of Psychiatric Rehabilitation, Shenzhen Mental Health Center/Shenzhen Kangning Hospital, Shenzhen, China (e-mail: qingzhu1108@126.com).}
\thanks{\textsuperscript{*}Corresponding authors: Zhen Liang and Yongjie Zhou.}}

\maketitle
\footnotetext[1]{\hspace{1mm}Equal contributions.}

\begin{abstract}
Non-suicidal self-injury (NSSI) is a serious threat to the physical and mental health of adolescents, significantly increasing the risk of suicide and attracting widespread public concern. Electroencephalography (EEG), as an objective tool for identifying brain disorders, holds great promise. However, extracting meaningful and reliable features from high-dimensional EEG data, especially by integrating spatiotemporal brain dynamics into informative representations, remains a major challenge. In this study, we introduce an advanced semi-supervised adversarial network, NSSI-Net, to effectively model EEG features related to NSSI. NSSI-Net consists of two key modules: a spatial-temporal feature extraction module and a multi-concept discriminator. In the spatial-temporal feature extraction module, an integrated 2D convolutional neural network (2D-CNN) and a bi-directional Gated Recurrent Unit (BiGRU) are used to capture both spatial and temporal dynamics in EEG data. In the multi-concept discriminator, signal, gender, domain, and disease levels are fully explored to extract meaningful EEG features, considering individual, demographic, disease variations across a diverse population. Based on self-collected NSSI data (n=114), the model's effectiveness and reliability are demonstrated, with a 5.44\% improvement in performance compared to existing machine learning and deep learning methods. This study advances the understanding and early diagnosis of NSSI in adolescents with depression, enabling timely intervention. The source code is available at \textit{https://github.com/Vesan-yws/NSSINet}.
\end{abstract}

\begin{IEEEkeywords}
Electroencephalography, NSSI, Decoding Model, Multi-Concept Discriminator, Domain Adaptation
\end{IEEEkeywords}

\section{Introduction}
\label{sec:introduction}
\IEEEPARstart{M}{ajor} depressive disorder (MDD) is a widespread chronic mental illness that manifests across all age groups \cite{xu2021scientometrics}. Adolescents, in particular, exhibit a notably high prevalence of MDD, with approximately 13\% of this population affected \cite{flores2023estimates}. Among those diagnosed with depression, nearly 50\% are reported to engage in non-suicidal self-injury (NSSI) behaviors \cite{swannell2014prevalence}. NSSI involves intentionally harming oneself, like cutting or burning, without the intent to die. It has drawn attention due to its connection to long-term psychological problems and increased risk of suicide \cite{cullen2020neural,levkovich2024non}. The rising occurrence of NSSI among adolescents with depression \cite{guan2024impact,zhang2024association} highlights an urgent need for effective tools and methods to understand, diagnose, and intervene in these behaviors. Early detection of NSSI behavior is absolutely vital for effective disease treatment and for preventing further severe harm. An automated detection approach not only aids in the management of underlying psychological issues but also significantly reduces the risk of the individual engaging in more dangerous or life-threatening actions in the future.

Electroencephalography (EEG) provides a direct method for reflecting neuronal dynamics originating in the central nervous system\cite{ye2024adaptive,liang2019unsupervised}, making it highly valuable in detecting abnormal brain activities. EEG has been widely used across various disease fields due to its non-invasive nature, high temporal resolution, and sensitivity to brain dynamics. For example, it plays a crucial role in epilepsy research \cite{smith2005eeg}, Alzheimer's Disease (AD) studies \cite{jeong2004eeg}, depression diagnosis and monitoring \cite{de2019depression}, autism spectrum disorder (ASD) analysis \cite{bosl2011eeg}, and schizophrenia research \cite{boutros2008status}. These applications underscore the versatility and importance of EEG in understanding and managing neurological and psychiatric conditions. In addition, the extraction of spatiotemporal features from EEG signals is crucial in revealing brain activity patterns related to diseases. For example, in seizure detection, EEGWaveNet developed a multiscale convolutional neural network (CNN) to extract spatiotemporal features \cite{thuwajit2021eegwavenet}, leading to significant improvements in detection accuracy. Liu \etal proposed a three-dimensional convolutional attention neural network (3DCANN) for EEG emotion recognition, integrating spatio-temporal feature extraction with an EEG channel attention module to effectively capture dynamic inter-channel relationships over time \cite{liu20213dcann}. Zhang \etal introduced two deep learning-based frameworks that utilize spatio-temporal preserving representations of raw EEG streams to accurately identify human intentions. These frameworks combine convolutional and recurrent neural networks, effectively exploring spatial and temporal information either in a cascade or parallel configuration \cite{zhang2019making}. Gao \etal developed a novel EEG-based spatial-temporal convolutional neural network (ESTCNN) designed to detect driver fatigue by leveraging the spatial-temporal structure of multichannel EEG signals \cite{gao2019eeg}. These studies highlight the effectiveness of spatiotemporal feature extraction in enhancing both the interpretability and performance of EEG-based disease studies.

However, in existing EEG-based disease studies, the inherent individual variability in EEG signals and the heavy reliance on labeling information present considerable challenges. This individual variability, which can arise from differences in brain anatomy, age, gender, and even the state of disease, often leads to inconsistencies in data interpretation and can significantly impact the generalization of intelligent models. The dependence on accurate and extensive labeling requires large, well-annotated datasets that are often time-consuming and costly to obtain. Moreover, labeling is subject to human error and bias, which can introduce noise into the training process, leading to models that may perform well on specific datasets but fail to generalize effectively to new, unseen data. To address the challenges present in existing EEG-based disease studies and to develop a robust, clinically applicable method for NSSI detection across diverse patient populations, we propose a novel multi-concept Generative Adversarial Network (GAN) framework, named NSSI-Net. This framework is developed using a semi-supervised learning approach, allowing it to effectively handle the variability in EEG signals and reduce dependency on extensive labeled data. The main contributions of this study are summarized as follows.

\begin{itemize}
    \item We propose a novel framework, NSSI-Net, which integrates a multi-concept discriminator architecture. This architecture incorporates discriminators at the signal, gender, domain, and disease levels, taking into account variations in individual and demographic characteristics and enabling the efficient extraction of spatiotemporal EEG features across a diverse population.
    \item NSSI-Net is designed to operate within a semi-supervised learning paradigm, reducing the model’s reliance on manually labeled data. It not only mitigates the challenges associated with obtaining large, fully labeled datasets but also enhances the model’s generalizability across different patient populations.
    \item The proposed NSSI-Net is validated using a self-constructed EEG dataset, consisting of recordings from 114 adolescents diagnosed with depression. Extensive experimental results demonstrate the efficiency and effectiveness of NSSI-Net in accurately detecting NSSI behaviors, establishing its potential for clinical application in diverse settings.
\end{itemize}

\begin{figure*}[h]
\centering
\includegraphics[width=1\textwidth]{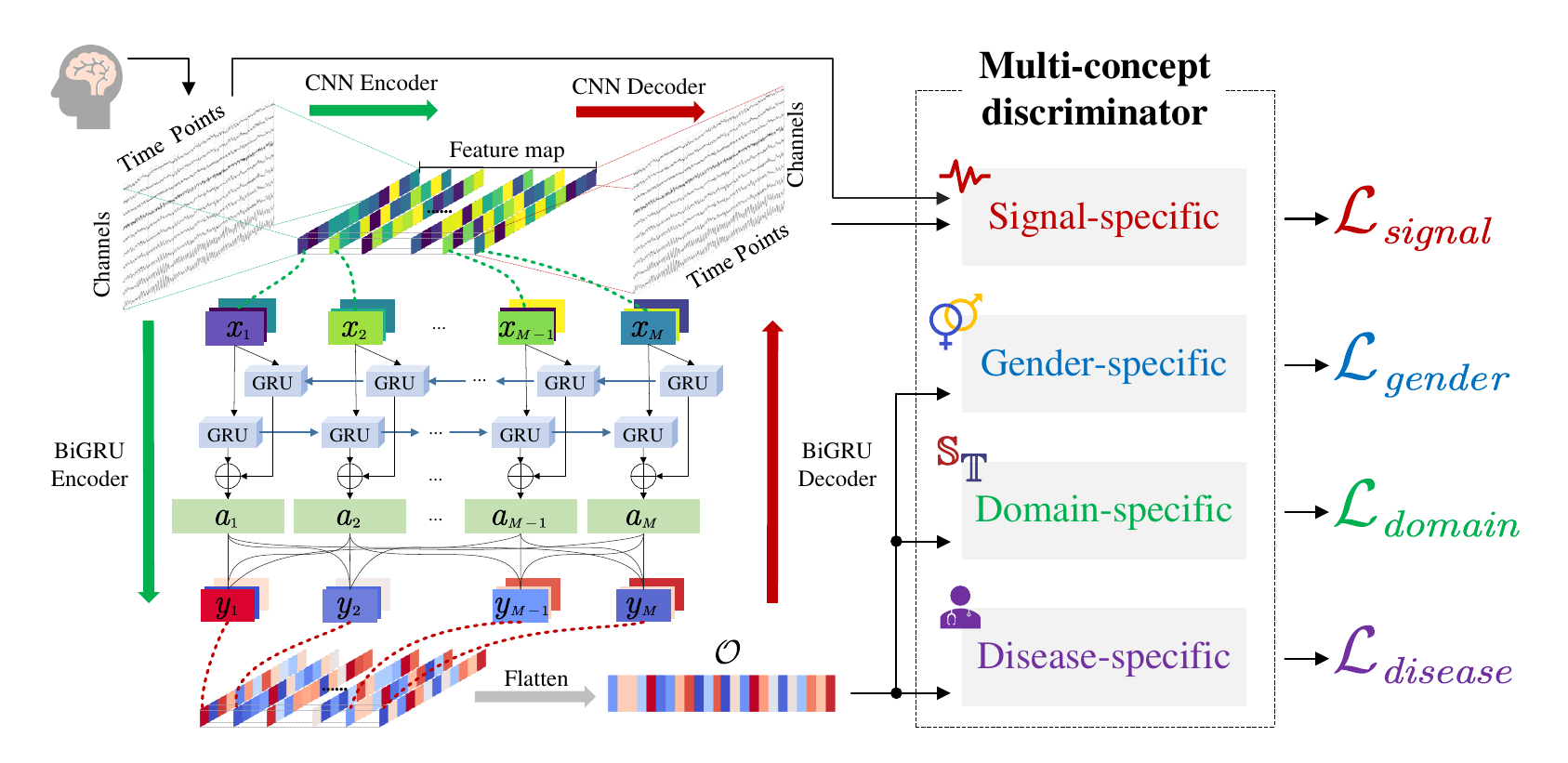}
\caption{The proposed framework consists of two primary components: a spatial-temporal feature extraction module and a multi-concept discriminator. The feature extraction module leverages an encoder-decoder structure incorporating CNN and BiGRU to capture spatiotemporal features from EEG signals. The multi-concept discriminator involves four specific-designed discriminators: (1) Signal-specific discriminator distinguishes between real and generated EEG signals to enhance feature validity; (2) Gender-specific discriminator differentiates EEG patterns across genders to extract more generalized features; (3) Domain-specific discriminator aligns features from labeled source, unlabeled source, and target domains to reduce domain discrepancies; and (4) Disease-specific discriminator identifies EEG characteristics distinguishing adolescents with depression who engage in NSSI from those who do not.}
\label{fig:pipeline}
\end{figure*}

\section{RELATED WORK}
\subsection{Neural Mechanisms in Depressed Adolescents with NSSI}
NSSI in adolescents with MDD has garnered significant research interest due to its rising prevalence and associated long-term psychological consequences. The neural underpinnings of NSSI in depressed youth are complex, involving both structural and functional alterations in brain regions critical for emotion regulation, impulse control, and self-referential processing. Recent neuroimaging studies, utilizing a range of techniques including EEG, magnetic resonance imaging (MRI), functional MRI (fMRI) and diffusion MRI (dMRI), have shed light on these neural mechanisms. 
Iznak \etal \cite{iznak2021eeg} utilized EEG identified notable differences in brain activity between depressive female adolescents with suicidal and non-suicidal auto-aggressive behaviors, particularly highlighting disruptions in regions involved in emotional regulation and impulse control. Auerbach \etal \cite{auerbach2021neural} reviewed MRI studies on NSSI in youth, identifying significant brain alterations. The research found reduced volume in the anterior cingulate cortex and other regions linked to emotion regulation. Functional MRI also showed altered frontolimbic connectivity and blunted striatal activation in adolescents with NSSI. Melinda \etal \cite{westlund2020white} used diffusion MRI to identify significant white matter microstructural deficits in adolescents and young adults with NSSI compared to healthy controls. These deficits may contribute to an increased vulnerability to maladaptive coping strategies like NSSI, with lower white matter integrity being linked to longer NSSI duration and higher impulsivity. The above studies collectively suggest that NSSI behaviors in depressed adolescents are driven by intricate neural mechanisms are highly related to the abnormalities in neural patterns. An efficient and reliable NSSI detection method using neural signals could reduce the occurrence and recurrence of NSSI, thereby preventing the escalation to more severe outcomes such as suicidal behavior in real life.

\subsection{EEG Studies Exploring NSSI in Depressed Adolescents}
In recent years, researchers have increasingly focused on utilizing EEG to explore the neural underpinnings of NSSI. In 2021, a study highlighted significant differences in EEG frequency and spatial components between depressive female adolescents who had attempted suicide and those who engaged in NSSI, demonstrating the potential of EEG as a diagnostic tool to distinguish between these two groups \cite{iznak2021differences}. Moreover, subsequent research has revealed notable alterations in emotion-related EEG components and brain lateralization in response to negative emotional stimuli among adolescents with NSSI, suggesting that these neural changes could be crucial in understanding the emotional dysregulation associated with NSSI \cite{zhao2023changes}. Further investigations have explored the effects of therapeutic interventions on EEG patterns in NSSI patients. For example, Zhao \etal 's work observed significant changes in EEG microstate patterns before and after repetitive transcranial magnetic stimulation (rTMS) treatment, indicating that EEG can not only reflects the abnormal neural activity related to NSSI but also monitors the impact of therapeutic interventions \cite{zhao2023changes2}. These findings underscore the importance of EEG as a non-invasive, real-time measure that can provide insights into the neural dynamics associated with NSSI, potentially leading to more targeted and effective treatments.

On the other hand, machine learning and deep learning approaches have also been investigated for analyzing EEG signals to predict NSSI behaviors. For example, Marti \etal used classification trees model that leverages EEG features to identify young adults at risk for NSSI \cite{marti2022machine}. Kim \etal developed a graph theory-based model utilizing resting-state EEG to analyze cortical functional networks and classify patients with MDD into those who have attempted suicide and those with suicidal ideation \cite{kim2024differentiation}. Kentopp \etal developed an adaptive transfer learning-based EEG classification model specifically designed to classify EEG signals of adolescents with and without a history of NSSI \cite{kentopp2021deep}. However, these approaches often depend heavily on prior knowledge for model parameter adjustment and require comprehensive label information for fully supervised learning. This reliance poses challenges in clinical settings, where data may be sparse, labels may be incomplete, and manual intervention in model tuning is impractical. These existing methods may fall short of meeting the complex demands of clinical practice. Therefore, there is a need for a more robust and adaptable end-to-end semi-supervised learning model capable of effectively addressing the complexities of clinical data.

\subsection{Deep Learning for Feature Extraction from High-Dimensional EEG Data}
Efficient feature extraction from high-dimensional EEG signals is essential for unraveling the complexities of brain functions and advancing accurate computational modeling\cite{ye2024semi}. Traditional machine learning methods often struggle to capture the inherent and dynamic nature of EEG signals \cite{marti2022machine}. Deep learning methods offer the capability to automatically identify and characterize informative spatio-temporal features within EEG data\cite{zhou2024eegmatch}, which not only enhance the precision of feature extraction but also allow for the dynamic adaptation to the non-linear and non-stationary characteristics of brain activity \cite{zhang2018spatial, li2021novel,lew2020eeg}. For example, a three-dimensional convolutional neural network (3D-CNN) was introduced, which processed EEG data as volumetric sequences and enhanced feature representation capabilities \cite{cho2020spatio}. Considering the temporal dynamics in EEG series, a hierarchical spatial-temporal neural network was proposed to analyze the transition from regional to global brain activity, demonstrating the efficacy of deep learning in capturing detailed spatiotemporal information \cite{li2019regional}. Besides of supervised learning, Liang \etal introduced a deep unsupervised autoencoder (AE) model (EEGFuseNet) designed to automatically characterize the interplay between spatial and temporal information in EEG data \cite{liang2021eegfusenet}. The latent features extracted by EEGFuseNet showed great potential for generalization across various applications. These works have underscored the transformative potential of deep learning in EEG research, which provide a more comprehensive, accurate and reliable understanding of brain activities.

\subsection{Semi-Supervised Learning in EEG Modeling}
Current deep learning approaches in EEG analysis are predominantly dependent on fully labeled datasets. However, in clinical environments, acquiring such well-annotated data is both challenging and expensive. Semi-supervised learning frameworks present a promising solution to this problem by enabling the use of both labeled and unlabeled data. Preliminary investigations into semi-supervised learning have shown encouraging results in various EEG studies. For example, a semi-supervised clustering method was proposed to effectively group EEG signals based on their underlying patterns, even with minimal labeled data \cite{dan2021possibilistic}. Further, Zhang \etal introduced a pairwise alignment of representations in semi-supervised EEG learning (PARSE), which enhances the consistency of feature representations across different data domains by aligning similar features from labeled and unlabeled data \cite{zhang2022parse}. Recently, Ye \etal proposed a semi-supervised dual-stream self-attentive adversarial graph contrastive learning model to effectively leverage limited labeled data and large amounts of unlabeled data for cross-subject EEG-based emotion recognition \cite{ye2024semi}. These methods successfully employed semi-supervised techniques to classify neurological conditions with a reduced amount of labeled data, achieving comparable accuracy to fully supervised models. This indicates the potential of semi-supervised learning in advancing EEG-based diagnostics and interventions, especially in resource-constrained clinical settings.

\section{PARTICIPANTS AND DATA}
\subsection{Participants}
We recruited a total of 114 adolescent patients, aged 13 to 18 years, from Shenzhen Kangning Hospital, all of whom had at least five years of education. These patients were diagnosed with MDD based on DSM-5 criteria and were further screened using the Mini-International Neuropsychiatric Interview (MINI) to exclude other psychiatric disorders. The study received approval from the Research Ethics Committee of Shenzhen Kangning Hospital (Approval No: 2020-K021-04-1). Informed consent was obtained from all participants and their guardians prior to enrollment.

In this study, NSSI behavior was defined based on the following criteria: (1) intentional self-inflicted injury to body tissue without suicidal intent within the past year; (2) the behavior served at least one of the following purposes: alleviating or relieving negative emotions, resolving interpersonal problems, or achieving a desired emotional state; (3) the self-injurious behavior was linked to at least one of the following: it occurred after interpersonal rejection or in response to negative emotions, there was an inability to resist the urge to self-harm, or thoughts of self-harm were present; (4) the behavior involved socially unacceptable actions, such as repeated cutting or picking at wounds. According to the NSSI criteria, the recruited patients were divided into two groups: Depression with NSSI (\textbf{DN+}), consisting of 77 subjects (65 females), and Depression without NSSI (\textbf{DN-}), consisting of 37 subjects (18 females).

\subsection{Data Acquisition and Preprocessing}
Each participant completed a total of 35 trials, with each trial involving the viewing of a natural image related to social pain. Each trial lasted 5 seconds. Throughout these trials, EEG data were simultaneously recorded using a 63-channel BrainAmp system, following the international 10-20 system. The data were captured at a sampling rate of 500 Hz.

A standard EEG preprocessing pipeline was applied to the raw EEG signals to remove artificial artifacts, including eye movements, muscle activity, and environmental noise. Following artifact correction, the signals were downsampled from 500 Hz to 384 Hz. This reduction not only decreases data dimensionality but also retains essential temporal, spatial, and frequency information, thereby enhancing the efficiency of subsequent processing and analysis \cite{liang2021eegfusenet}. To increase the sample size for modeling, each 5-second trial was further segmented into multiple 1-second samples. As a result, the final data format can be represented as 114 participants $\times$ 35 trials $\times$ 5 seconds $\times$ 63 channels $\times$ 384 sampling points.

\begin{figure*}[]
	\centering
	  \includegraphics[width=1\textwidth]{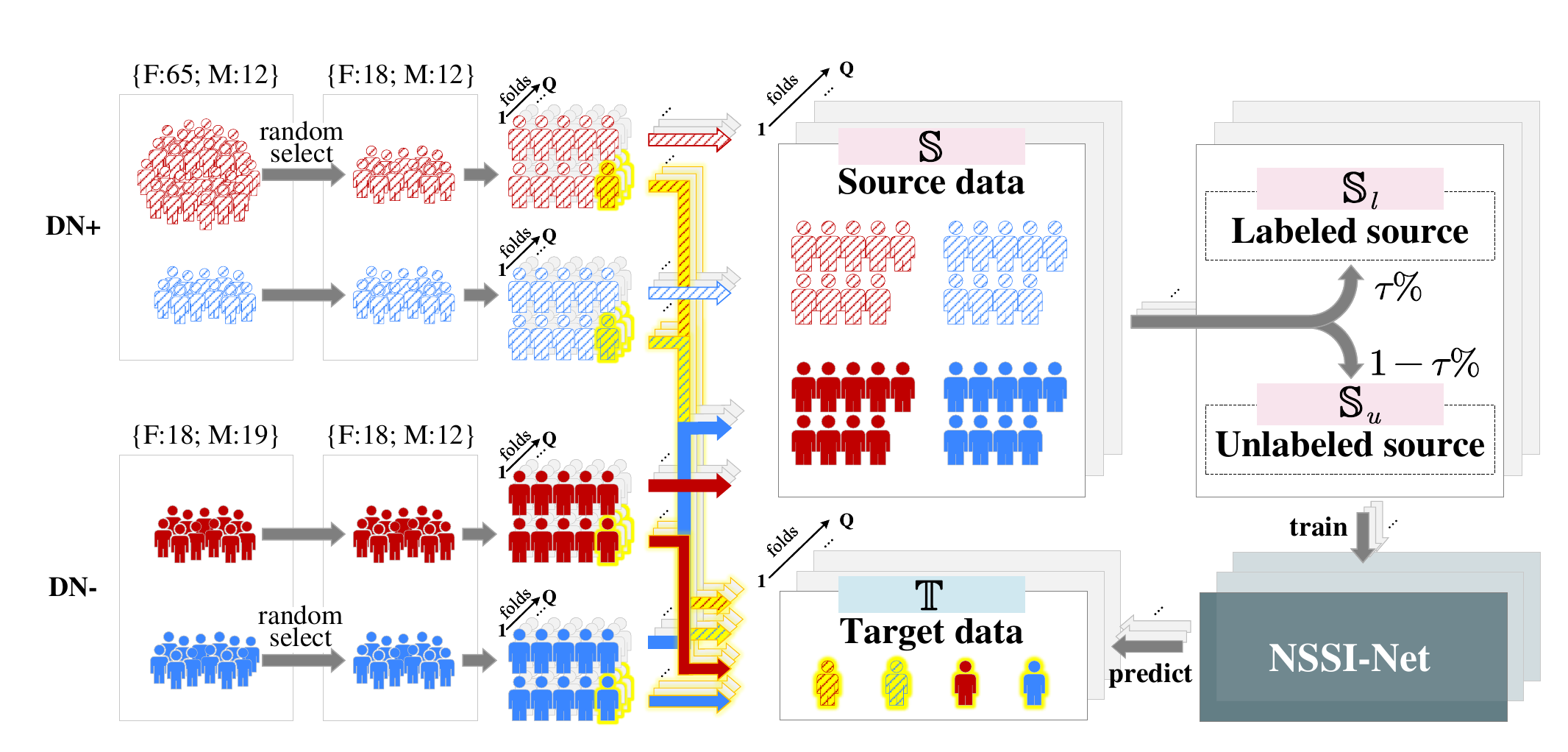}
	\caption{The sampling process strategy, considering the data balance and gender distribution in semi-supervised cross-subject cross-validation.}
 \label{fig:sampling}
\end{figure*}

\section{METHODOLOGY}
This paper presents a novel semi-supervised learning framework (NSSI-Net), which incorporates a multi-concept GAN to characterize informative EEG patterns related to NSSI. NSSI-Net tackles the clinical challenge of label scarcity in EEG data through two key modules: an encoder-decoder-based feature extraction module and a multi-concept discriminator module, as illustrated in Fig. \ref{fig:pipeline}. \textbf{Encoder-Decoder-based Feature Extraction Module}. The encoder-decoder architecture is constructed using CNN and BiGRU to efficiently capture both spatial and temporal representations of NSSI-related brain activities. This module enables the extraction of meaningful patterns from the EEG signals by modeling their spatial dependencies and temporal dynamics. \textbf{Multi-Concept Discriminator Module.} The extracted EEG features are further refined to enhance the robustness and generalizability of the representations. This refinement process operates across four concepts: signal, gender, domain, and disease. A detailed explanation of the multi-concept discriminators will be provided in Section \ref{sec:multiCon-Discriminator}.

\subsection{Input Data}
\label{sec:inputData}
Based on the collected data, we categorize it into two domains: the source domain $\mathbb{S}$ and the target domain $\mathbb{T}$. Within the semi-supervised learning framework, the source domain ($\mathbb{S}$) is further subdivided into a smaller labeled subset (denoted as $\mathbb{S}_l$) and a larger unlabeled subset (denoted as $\mathbb{S}_u$). Here, a division ratio, denoted as $\tau\%$, determines that $\tau\%$ of the source data forms the labeled subset ($\mathbb{S}_l$), while the remaining $(1-\tau\%)$ constitutes the unlabeled subset ($\mathbb{S}_u$).

Considering the data imbalance (NSSI: 77 patients, including 65 females and 12 males; non-NSSI: 37 patients, including 18 females and 19 males) and the gender distribution, a data sampling strategy is implemented across $\mathbb{S}_l$, $\mathbb{S}_u$, and $\mathbb{T}$. As illustrated in Fig. \ref{fig:sampling}, the sampling process for both the NSSI and non-NSSI groups is explained. For the NSSI group (denoted as DN+), 18 females are randomly selected from the 65 available candidates (red hatched), and all 12 males are included, ensuring equal gender representation within DN+. For the non-NSSI group (denoted as DN-), all 18 females (solid blue) are included, along with 12 randomly selected males from the 19 available candidates, achieving a balanced gender distribution comparable to the NSSI group. This sampling process results in a dataset of 30 NSSI patients (DN+) and 30 non-NSSI patients (DN-), balanced across both condition and gender.

\subsection{Spatial-Temporal Feature Extraction}
\label{sec:Feature_Extraction_via_Encoder-Decoder_Structure}
To effectively extract spatial and temporal features from input EEG signals, we design a hybrid neural network that combines CNNs with GRUs. This architecture captures interactions both among different brain regions and across various time points, enhancing the representation of complex EEG dynamics.

Inspired by the EEGNet structure \cite{lawhern2018eegnet}, we develop a deep encoder-decoder network based on CNNs. The CNN-based encoder for spatial information extraction is shown in Fig. \ref{fig:pipeline}, represented by the green horizontal arrow. The input EEG sample is denoted as $X \in \mathbb{R}^{C\times P}$, where $C$ is the number of EEG channels and $P$ represents the number of sampling points per second. The CNN-based feature extractor adopts 2D convolution to map the input EEG samples into a more expressive feature space, providing a compact yet comprehensive representation of the spatial dynamics of the EEG signals. The obtained feature map is represented as $\mathcal{X} = (x_1, ..., x_M) \in \mathbb{R}^{M\times 1\times F}$, where $M$ is the number of feature maps and $F$ is the feature dimensionality.

To further uncover the relationships of the brain patterns over time and encapsulate the key temporal dynamics of the entire EEG signal sequence, a BiGRU based encoder-decoder structure is designed. The BiGRU-based encoder for temporal information extraction is illustrated in Fig. \ref{fig:pipeline}, represented by the green vertical arrow. In our implementation, we use a BiGRU to process the sequence of feature vectors generated by the CNN encoder. The forward and backward passes of the BiGRU produce corresponding hidden states, which are concatenated to form the deep feature representation vector $\mathcal{A} = (a_1, ..., a_M) \in \mathbb{R}^{M\times 1\times 2F}$. Once $\mathcal{A}$ is obtained from the concatenated hidden states of the BiGRU, it is fed into a Multi-Layer Perceptron (MLP) to derive $\mathcal{Y} = (y_1, ..., y_M) \in \mathbb{R}^{M\times 1\times F}$. This MLP transforms the feature representation into a more refined feature space, capturing non-linear relationships between the elements of $\mathcal{A}$. Following this, a flatten operation is applied to $\mathcal{Y}$, resulting in the output vector $\mathcal{O} \in \mathbb{R}^{1\times MF}$. The output $\mathcal{O}$ is a rich representation that integrates both spatial features (extracted by the CNN layers) and temporal features (learned through the recurrent layers), providing a comprehensive understanding of the EEG signal's dynamics.

\subsection{Multi-Concept Discriminator}
\label{sec:multiCon-Discriminator}
Traditional encoder-decoder networks, though easy to train, often produce lower-quality features and limit the effectiveness in complex tasks \cite{akbari2018semi}. In contrast, GANs have shown great potential in generating high-quality features from time-series data \cite{makhzani2015adversarial,chen2018unsupervised,sahu2018adversarial}. To enhance the robustness and accuracy of EEG signal classification, particularly when dealing with the complexities of non-stationary time-series data, we develop a multi-concept discriminator as part of our hybrid CNN-BiGRU encoder-decoder structure. This multi-concept discriminator is designed to enhance the model's quality and generalizability by addressing multiple aspects of the data, covering signal quality, gender differences, domain variability, and disease-specific characteristics.

\subsubsection{Signal-Specific Discriminator}

To identify the quality of signals generated by the CNN-BiGRU encoder-decoder architecture, a signal-specific discriminator is designed. This discriminator aims to enhance the feature representation capability by distinguishing between generated signals and authentic ones. By providing adversarial feedback, the discriminator plays a crucial role in boosting the feature characterization capacity of the entire model, ensuring that the generated signals are of high quality and effectively represent the desired characteristics. Specifically, the generator $G$ extracts latent features $\mathcal{Y}$ from the input EEG signals $X$ and then reconstructs the EEG signals $G(X)$. The signal-specific discriminator is $D$. The loss function of the signal-specific discriminator is given as
\begin{equation}
\label{Eq:l_signal}
\mathcal{L}_{signal} = \text{arg}\min_{G}\max_{D}(\mathcal{L}_{GAN}(G,D) + \lambda\mathcal{L}_{1}(G)).
\end{equation}
$\mathcal{L}_{GAN}$ is the GAN loss, defined as
\begin{equation}
\label{Eq:l_gan}
\mathcal{L}_{GAN}(G,D) = {\mathbb{E}_{X}}[\log{D(X)}] + {\mathbb{E}_{X}}[\log{1-D(G(X))}].
\end{equation}
$\mathcal{L}_{1}$ is reconstruction loss, defined as
\begin{equation}
\label{Eq:l_G}
\mathcal{L}_{1}(G) = \|X - G(X)\|_2^2.
\end{equation}
$\lambda$ is a hyperparameter balancing the GAN loss $\mathcal{L}_{GAN}$ and the reconstruction loss $\mathcal{L}_{1}$, encouraging the generation of high-quality EEG signals.

\subsubsection{Gender-Specific Discriminator}
Considering the gender bias present in NSSI patients and the variations in EEG representations between females and males, a gender-specific discriminator is designed to capture and leverage gender-specific EEG features. By integrating gender-specific discriminator, the model not only accounts for the gender-based differences but also improves its overall classification accuracy, ensuring that the subtleties in EEG signals attributed to gender are adequately represented and utilized for better prediction outcomes. The loss function of the gender-specific discriminator is given as

\begin{equation}
\label{Eq:l_gender}
\mathcal{L}_{gender}\left(\theta_{G}, \theta_{e}\right) = -\sum_{i=0}^{N_{\mathbb{M}}} \log e\left(\mathcal{O}_{i}^{\mathbb{M}}\right) - \sum_{i=0}^{N_{\mathbb{F}}} \log \left(1-e\left(\mathcal{O}_{i}^{\mathbb{F}}\right)\right).
\end{equation}
Here, $\theta_{G}$ represents the parameters of the feature extractor (generator $G$). $\theta_{e}$ refers to the parameters of the gender classifier $e(\cdot)$. $\mathcal{O}$ refers to the features extracted as described in Section \ref{sec:Feature_Extraction_via_Encoder-Decoder_Structure}, where $\mathcal{O}^{\mathbb{M}}$ and $\mathcal{O}^{\mathbb{F}}$ represent the features of male and female samples, respectively. $N_{\mathbb{M}}$ and $N_{\mathbb{F}}$ denote the number of male and female samples in each batch, respectively. This loss function is designed to ensure that the model effectively captures gender-specific EEG features, which are crucial, particularly in addressing the gender bias present in the data.

\subsubsection{Domain-Specific Discriminator}
Due to the inherent individual differences in EEG signals, a domain-specific discriminator is designed to align the feature distributions across the labeled source domain ($\mathbb{S}_l$), the unlabeled source domain ($\mathbb{S}_u$), and the target domain ($\mathbb{T}$). This discriminator aims to reduce the variability among these domains by ensuring that the extracted features from different sources are more consistent. By aligning these distributions, the domain-specific discriminator helps in minimizing the domain gap, thereby enhancing the model’s ability to generalize effectively to new and unseen data. The loss function of the domain-specific discriminator is given as
\begin{equation}
\label{Eq:l_disc}
\mathcal{L}_{disc}\left(\theta_{G}, \theta_{d}\right) = -\sum_{i=0}^{N} p\left(X_i\right) \log d(\mathcal{O}_i).
\end{equation}
Here, $\theta_{d}$ refers to the parameters of the domain classifier $d(\cdot)$. $p(X)$ is the domain label indicating whether $X$ is from the labeled source, unlabeled source, or target domain. $N$ denotes the number of samples in each batch. This loss function encourages the feature extractor to generate domain-invariant features, by minimizing the ability of the domain classifier to distinguish among domains. As a result, the model learns to produce features that are robust across different domains, improving cross-subject generalization.

\subsubsection{Disease-Specific Discriminator}
To effectively capture the differential EEG features between DN+ and DN-, a disease-specific discriminator is designed. This discriminator plays a crucial role in distinguishing patients based on their EEG patterns, ensuring that the unique characteristics associated with each condition are accurately identified. The loss function of the disease-specific discriminator is given as
\begin{equation}
\label{Eq:l_disease}
\mathcal{L}_{disease} = -\sum_{i=0}^{N_{\mathbb{S}_l}} (y_i \log \sigma(f(\mathcal{O}_i)) + (1-y_i) \log (1-\sigma(f(\mathcal{O}_i)))),
\end{equation}
where $y$ represents the binary labels indicating the presence or absence of NSSI (DN+ or DN-). $N_{\mathbb{S}_l}$ denotes the number of samples in a batch from the labeled source domain, $f(\cdot)$ is the disease classifier, and $\sigma$ is the sigmoid activation function. This loss function aims to drive the model to accurately classify patients based on their EEG features, enhancing its predictive capability for identifying NSSI.

\subsubsection{Overall Loss Function}
The overall objective of the multi-concept discriminator is formulated as a weighted combination of the individual loss functions as

\begin{equation}
\label{Eq:l_total}
\mathcal{L} = \alpha \mathcal{L}_{signal} + \beta \mathcal{L}_{gender} + \delta \mathcal{L}_{disc} + \theta \mathcal{L}_{disease},
\end{equation}
where $\alpha$, $\beta$, $\delta$ and $\theta$ are weighting factors for the corresponding loss terms. The overall loss function ensures that the model learns robust, multi-dimensional EEG feature representations that are invariant to signal noise, sensitive to gender differences, adaptable across domains, and effective in classifying NSSI behaviors. By jointly optimizing these objectives, the model is better equipped to handle the complexities involved in EEG signal analysis for studies on depression and NSSI.

\section{Experimental Results}

\subsection{Implementation details and model setting}
A detailed summary of the model architecture, including the encoder and decoder configurations, is provided in Table \ref{tab:model_structure}. The CNN encoder begins with an initial convolutional layer that employs 16 filters and a kernel size of 1 $\times \frac{\text{input\_size}}{2} + 1$. This layer is followed by a depthwise convolution with 32 filters. Subsequent separable convolutions further refine these features, using kernel sizes of 1 $\times \frac{\text{input\_size}}{8} + 1$. The BiGRU encoder is bidirectional, with each direction containing 128 units per layer, spread across 2 layers. Following the BiGRU layers, fully connected layers are employed to condense the feature representation, which systematically reduce and refine the feature dimensionality before passing it to the final classification and adversarial networks. The adversarial discriminators incorporate fully connected layers with ReLU activation functions.

For model training, the RMSprop optimizer is used with a learning rate of 1e-3. The mini-batch size is set to 48, and L2 regularization with a coefficient of 1e-5 is applied to mitigate overfitting. Dropout is incorporated at a rate of 0.25 in the fully connected layers to improve generalization. The model is validated using a semi-supervised cross-subject ten-fold cross-validation. The division of labeled source data and unlabeled source data is based on subjects rather than data samples. Specifically, based on the training data, 75\% of the subjects from the source domain are randomly assigned as labeled source data, and the rest 25\% subjects are unlabeled source data. All parameters are randomly initialized, and training is conducted on a single GPU to ensure efficiency and consistency throughout the process.

\begin{table}[]
\caption{Summary of model architecture and configurations.}
\label{tab:model_structure}
\centering
\setlength{\tabcolsep}{5pt} 
\begin{tabular}{lccc}
\toprule
Layer (type:depth-idx) & Kernel Shape & Output Shape & Param\# \\\midrule
Input & -- & [384, 1, 63, 384] & --\\
Conv2d: 1-1 & [1, 193] & [384, 16, 63, 384] & 3,104 \\
BatchNorm2d: 1-2 & -- & [384, 16, 63, 384] & 32 \\
Conv2d: 1-3 & [63, 1] & [384, 32, 1, 384] & 32,288 \\
BatchNorm2d: 1-4 & -- & [384, 32, 1, 384] & 64 \\
MaxPool2d: 1-5 & [1, 4] & [384, 32, 1, 96] & -- \\
Dropout: 1-6 & -- & [384, 32, 1, 96] & -- \\
Conv2d: 1-7 & [1, 49] & [384, 32, 1, 96] & 1,600 \\
Conv2d: 1-8 & [1, 1] & [384, 16, 1, 96] & 528 \\
BatchNorm2d: 1-9 & -- & [384, 16, 1, 96] & 32 \\
MaxPool2d: 1-10 & [1, 8] & [384, 16, 1, 12] & -- \\
Linear: 1-11 & -- & [384, 12, 16] & 272 \\
GRU: 1-12 & -- & [384, 12, 32] & 3,264 \\
Linear: 1-13 & -- & [384, 12, 16] & 528 \\
Linear: 1-14 & -- & [384, 12, 32] & 544 \\
GRU: 1-15 & -- & [384, 12, 32] & 4,800 \\
Linear: 1-16 & -- & [384,12,16] & 528\\
MaxUnpool2d: 1-17 & [1, 8] & [384, 16, 1, 96] & -- \\
ConvTranspose2d: 1-18 & [1, 1] & [384, 32, 1, 96] & 544 \\
ConvTranspose2d: 1-19 & [1, 49] & [384, 32, 1, 96] & 1,600 \\
BatchNorm2d: 1-20 & -- & [384, 32, 1, 96] & 64 \\
Dropout: 1-21 & -- & [384, 32, 1, 96] & -- \\
MaxUnpool2d: 1-22 & [1, 4] & [384, 32, 1, 384] & -- \\
ConvTranspose2d: 1-23 & [63, 1] & [384, 16, 63, 384] & 32,272 \\
BatchNorm2d: 1-24 & -- & [384, 16, 63, 384] & 32 \\
ConvTranspose2d: 1-25 & [1, 193] & [384, 1, 63, 384] & 3,089 \\
\midrule
\multicolumn{4}{l}{Total params: 164,807} \\
\multicolumn{4}{l}{Trainable params: 164,807} \\
\multicolumn{4}{l}{Non-trainable params: 0} \\
\multicolumn{4}{l}{Total mult-adds (G): 362.29} \\
\midrule
\multicolumn{4}{l}{Input size (MB): 37.16} \\
\multicolumn{4}{l}{Forward/backward pass size (MB): 4958.66} \\
\multicolumn{4}{l}{Params size (MB): 0.34} \\
\multicolumn{4}{l}{Estimated Total Size (MB): 4996.16} \\
\bottomrule
\end{tabular}
\end{table}

\subsection{Model Results under Semi-Supervised Learning}
We compare the proposed NSSI-Net with various machine learning and deep learning models using a semi-supervised cross-subject ten-fold cross-validation protocol. 
In this framework, the labeled source domain $\mathbb{S}_l$ contributes to the optimization of all four discriminators: $\mathcal{L}_{signal}$, $\mathcal{L}_{gender}$, $\mathcal{L}_{domain}$ and $\mathcal{L}_{disease}$. The unlabeled source domain $\mathbb{S}_u$ and the target domain $\mathbb{T}$ contribute to the optimization of three discriminators: $\mathcal{L}_{signal}$, $\mathcal{L}_{gender}$, and $\mathcal{L}_{domain}$. Here, $\mathcal{L}_{disease}$ is trained in a supervised manner using the labeled data from $\mathbb{S}_l$. $\mathcal{L}_{signal}$, $\mathcal{L}_{gender}$, and $\mathcal{L}_{domain}$ leverage self-supervised learning to extract meaningful feature representations without requiring label information. As shown in Table \ref{tab:nssisemicompare}, the proposed NSSI-Net achieves superior performance, with an accuracy of 70.00±13.90. This represents a significant improvement of 5.44\% compared to the best-performing model in the literature.

Specifically, the table could be divided into two parts: traditional machine learning methods and deep learning methods. 
Among traditional machine learning approaches, Adaboost \cite{2006Boost} achieves the highest accuracy at 57.85±06.57, followed closely by CORAL \cite{CORAL2016} with 57.07±03.45. Other traditional models, including SA\cite{SA2013}, RF\cite{Breiman2001RF}, SVM \cite{SVM1999}, TCA \cite{TCA2010}, KPCA \cite{KPCA1999}, and GFK \cite{GFK2012}, demonstrate comparatively lower performance, with accuracies ranging from 48.09\% to 56.87\%. On the other hand, the deep learning methods offer more competitive results. Among transfer learning approaches, DDC \cite{DDC2014} achieves an accuracy of 62.56±02.52, followed by DAN \cite{He2018DAN} at 61.46±03.86, DCORAL \cite{Dcoral2016}  at 60.91±02.95, and DANN \cite{ganin2016domain} at 58.20±02.80. Among existing methods, the best semi-supervised learning results are achieved by MixMatch \cite{berthelot2019mixmatch} with an accuracy of 64.56$\pm$04.46, followed by PARSE \cite{zhang2022parse} at 61.77$\pm$13.34. Through a comprehensive comparison with the literature, the proposed NSSI-Net proves to be the top-performing model, surpassing both traditional and deep learning methods. Its 5.44\% accuracy improvement over the second-best model (MixMatch \cite{berthelot2019mixmatch}) highlights its effectiveness in capturing the complex patterns associated with NSSI behaviors in adolescents, establishing it as a promising tool for early detection and intervention.

\begin{table}[]
\begin{center}
\caption{The mean accuracies (\%) and standard deviations (\%) of various traditional machine learning and deep learning methods for NSSI detection using semi-supervised cross-subject ten-fold cross-subject cross-validation. Models reproduced by the authors are indicated with an asterisk (*).}
\label{tab:nssisemicompare}
\begin{tabular}{lc|lc} 
\toprule
Methods   & Accuracy   & Methods    & Accuracy    \\ 
\midrule
\multicolumn{4}{c}{\textbf{\textit{Traditional machine learning methods}}}      \\ 
\midrule
SVM*\cite{SVM1999}   &56.87$\pm$04.96 & TCA*\cite{TCA2010}   &53.02$\pm$04.41  \\
SA*\cite{SA2013}     &48.09$\pm$06.85 & KPCA*\cite{KPCA1999} &51.28$\pm$05.96  \\
RF*\cite{Breiman2001RF}  &55.39$\pm$07.04 & Adaboost*\cite{2006Boost} &57.85$\pm$06.57 \\
CORAL*\cite{CORAL2016}   &57.07$\pm$03.45 & GFK*\cite{GFK2012}   &56.33$\pm$06.52 \\
\midrule
\multicolumn{4}{c}{\textit{\textbf{\textbf{Deep learning methods}}}}            \\ 
\midrule
DAN*\cite{He2018DAN} &61.46$\pm$03.86 & DANN*\cite{ganin2016domain}  &58.20$\pm$02.80  \\
DCORAL*\cite{Dcoral2016}  & 60.91$\pm$02.95 & DDC*\cite{DDC2014}  & 62.56$\pm$02.52  \\
MixMatch*\cite{berthelot2019mixmatch} & 64.56$\pm$04.46 & PARSE*\cite{zhang2022parse} & 61.77$\pm$13.34\\
\midrule
\multicolumn{2}{l}{\textbf{NSSI-Net}} & \multicolumn{2}{r}{\textbf{70.00$\pm$13.90} \textcolor{red}{\textbf{(+5.44)}}} \\
\bottomrule
\end{tabular}
\end{center}
\end{table}

\subsection{Analysis of Confusion Matrices}
\begin{figure*}[h]
\begin{center}
\includegraphics[width=0.9\textwidth]{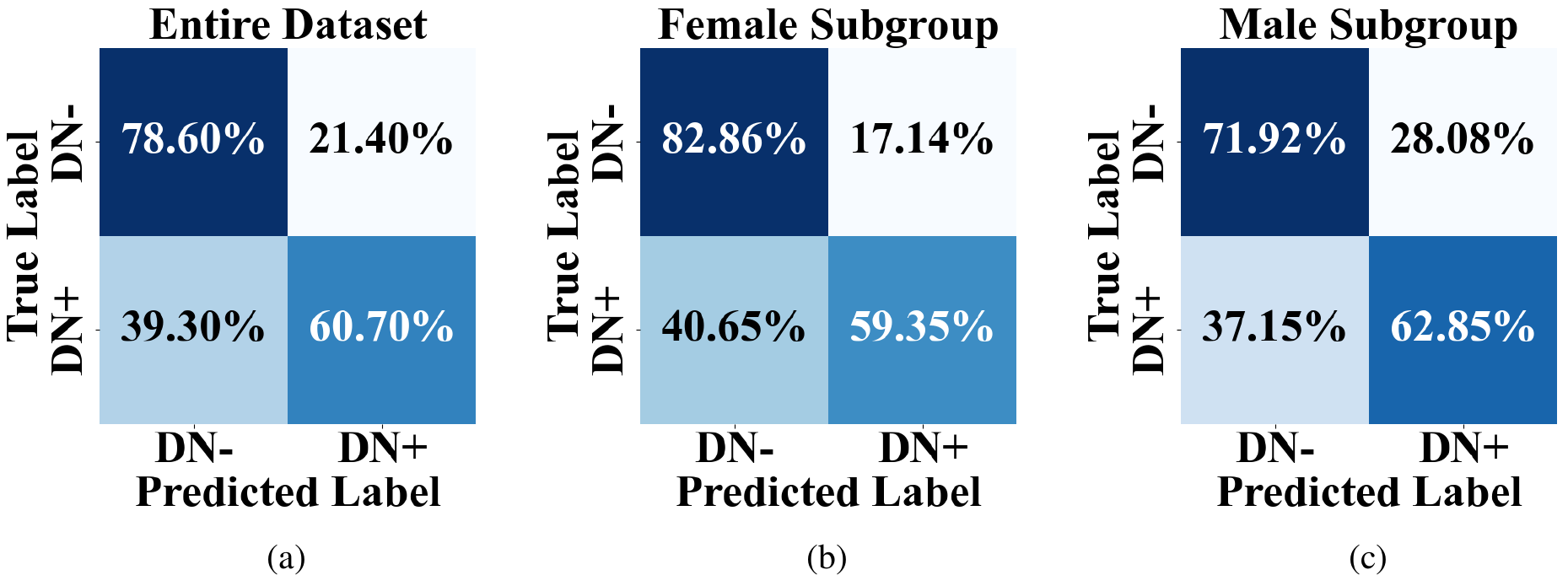}
\caption{The confusion matrices for different groups, where DN+ is considered as the positive class and DN- is considered as the negative class. (a) The overall confusion matrix. (b) The confusion matrix specific to the female subgroup. (c) The confusion matrix specific to the male subgroup. Each matrix displays the proportion of true positive (TP), true negative (TN), false positive (FP), and false negative (FN) predictions, with darker colors indicating higher percentages.}
\label{fig:confusion_matrix}
\end{center}
\end{figure*}

To further evaluate model performance across different subgroups and identify any biases or inconsistencies in the prediction outcomes, we analyze the confusion matrices for three demographic groups: the entire dataset, the female subgroup, and the male subgroup. As shown in Fig. \ref{fig:confusion_matrix}, we consider DN+ as the positive class and DN- as the negative class.

The overall confusion matrix reflects the model's performance across the entire dataset. The true negative (TN) rate is 78.60\%, and the true positive (TP) rate is 60.70\%, indicating that the model is reasonably effective at distinguishing between DN+ and DN- cases. However, the false positive (FP) rate of 21.40\% and false negative (FN) rate of 39.30\% reveal limitations, particularly in identifying DN+ cases. The higher FN rate suggests difficulty in correctly identifying DN+ instances, potentially due to the inherent variability in EEG signals among individuals.

For the female subgroup, the confusion matrix shows a TN rate of 82.86\% and a TP rate of 59.35\%, indicating slightly better performance in correctly identifying DN- cases compared to DN+ cases. The FP rate is lower at 17.14\%, while the FN rate is 40.65\%. These results imply that the model is more accurate in predicting DN- in females, potentially due to the higher prevalence of NSSI in females, making the EEG patterns associated with DN- more distinguishable. However, the relatively high FN rate indicates considerable room for improvement in detecting DN+ cases in this subgroup.

For the male subgroup, the confusion matrix reveals a different pattern, with a lower TN rate of 71.92\% but a higher TP rate of 62.85\%. The FP rate is notably higher at 28.08\%, while the FN rate is lower at 37.15\%. This distribution suggests that the model is more sensitive to detecting DN+ cases in males but at the cost of a higher rate of false positives, indicating a potential gender-specific bias. This bias might be influenced by the generally lower prevalence of NSSI among males, making it more challenging for the model to distinguish between DN- and DN+ cases in this subgroup.

In summary, our current study reveals that the training of our gender-specific discriminator is likely constrained by two main factors: dataset imbalance and inherent biological and behavioral differences between genders. Due to the uneven gender distribution, the model may predominantly learn features from the female subgroup, which exhibits greater variance. Although we implement a data sampling strategy to minimize the impact of this imbalance, the lower variance in the male subgroup still leads to biased feature learning. Moreover, neurophysiological and behavioral markers associated with NSSI differ between genders. This variability makes it challenging for a uniform training process to fully capture the unique EEG patterns of each gender, thereby limiting the overall performance of the discriminator.

\section{Discussion and Conclusion}
To thoroughly assess the performance and robustness of the proposed NSSI-Net, we conduct a series of experiments to evaluate the contribution of each component within the model. We also examine the impact of various hyperparameter settings on its performance. Additionally, we discuss the revealing neural mechanisms observed in NSSI-Net. These experimental analyses provide a comprehensive understanding of the model's capabilities and help identify the factors most crucial for its success in detecting NSSI.

\subsection{Ablation Study}
In this section, we conduct an ablation study to evaluate the contribution of each discriminator in the proposed model. Table \ref{tab:ablation} presents the model results using semi-supervised cross-subject ten-fold cross-validation under different discriminator configurations. \textbf{(1) Signal-Specific Discriminator Contribution.} Removing the signal-specific discriminator reduces the accuracy to 65.47±10.22, highlighting its crucial role in extracting relevant signal-based features. \textbf{(2) Gender-Specific Discriminator Contribution.} Without the gender-specific discriminator, the accuracy drops slightly to 66.63±09.66, indicating that gender-specific features contribute to overall classification performance, though less significantly than signal-specific features. \textbf{(3) Domain-Specific Discriminator Contribution.} When the domain-specific discriminator is removed, the accuracy decreases to 64.30±09.88, showing the importance of domain adaptation in improving the model's performance. \textbf{(4) Comparison with Traditional Domain Discriminator.} Traditional domain adaptation methods are based on the DANN framework \cite{ganin2016domain}. In line with previous approaches \cite{li2019multisource, chen2021ms}, we treat both labeled and unlabeled source data as a unified domain for adaptation. Replacing the proposed discriminator setup with a traditional domain discriminator yields an accuracy of 67.17±12.99, which is lower than the proposed NSSI-Net's performance of 70.00±13.90, underscoring the effectiveness of the proposed multi-concept discriminator framework. The results demonstrate the importance of each discriminator in enhancing the model's performance, with the signal-specific and domain-specific discriminators having the most significant impact.

\begin{figure*}
\begin{center}
\includegraphics[width=1\textwidth]{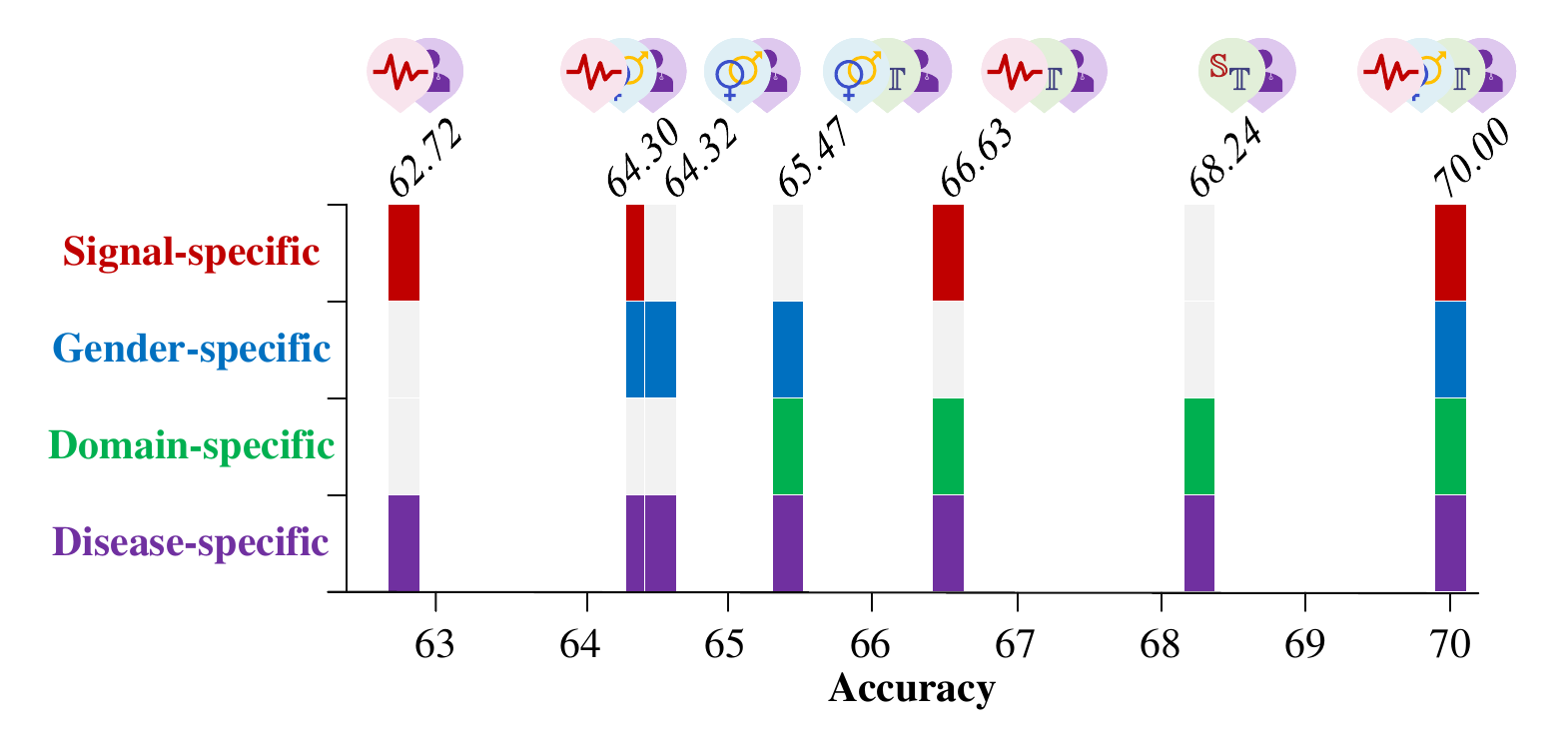}
\end{center}
\caption{The contribution of different model components (signal-specific, gender-specific, and domain-specific modules) towards the prediction accuracy of the model. Each vertical bar represents a specific combination strategy. Red, blue, green, and purple indicate the inclusion of signal-specific, gender-specific, domain-specific and disease-specific discriminator, respectively. Gray shows the absence. Combining more modules improves the model's accuracy, with the highest accuracy of 70.00\% achieved when all components are included.}
\label{fig:acc_contributions}
\end{figure*}

In addition to removing individual discriminators to evaluate the contribution of each component, we also explore different combinations of discriminators (signal-specific, gender-specific and domain-specific) to assess their collective impact on the overall prediction accuracy of the model. The model performance under different combinations are presented in Fig. \ref{fig:acc_contributions}. \textbf{(1) Signal + Disease-Specific.} When using only the signal-specific module, the model achieves a baseline accuracy of 62.72\%, highlighting the importance of capturing key signal-related features for distinguishing between classes. \textbf{(2) Signal + Gender + Disease-Specific.} Adding the gender-specific module alongside the signal-specific module raises the accuracy to 64.30\%, demonstrating the added value of incorporating gender-related information, as gender-specific EEG patterns contribute additional predictive power. \textbf{(3) Gender + Domain + Disease-Specific.} The combination of gender-specific and domain-specific modules yields an accuracy of 65.47\%. This shows that combining domain-related information with gender-specific features further improves prediction, indicating the significance of domain variations for classification. \textbf{(4) Signal + Gender + Domain + Disease-Specific.} The highest accuracy, 70.00\%, is achieved when all discriminators are combined. This demonstrates that each discriminator captures unique, complementary aspects of the data, and that their integration provides the most accurate predictions.

The above analyses show that the signal-specific module captures foundational features, while the gender-specific and domain-specific modules provide additional refinement by accounting for gender and domain-specific variations. Although individual modules are informative, their combined effect maximizes the model’s predictive performance.

\begin{table}[]
\begin{center}
\caption{The ablation study of the proposed NSSI-Net.}
\label{tab:ablation}
\begin{tabular}{l@{\hskip 52pt}c} 
\toprule
Methods   & Accuracy     \\ 
\midrule
Without signal-specific discriminator   &65.47$\pm$10.22  \\
Without gender-specific discriminator  &66.63$\pm$09.66  \\
Without domain-specific discriminator  &64.30$\pm$09.88   \\
With tradition domain discriminator  &67.17$\pm$12.99   \\
\midrule
\textbf{NSSI-Net} & \textbf{70.00$\pm$13.90}  \\
\bottomrule
\end{tabular}
\end{center}
\end{table}

\subsection{Evaluating Model Performance under Varying Labeled Source Data Ratios in Semi-Supervised Learning}

\begin{figure}
\begin{center}
\includegraphics[width=0.5\textwidth]{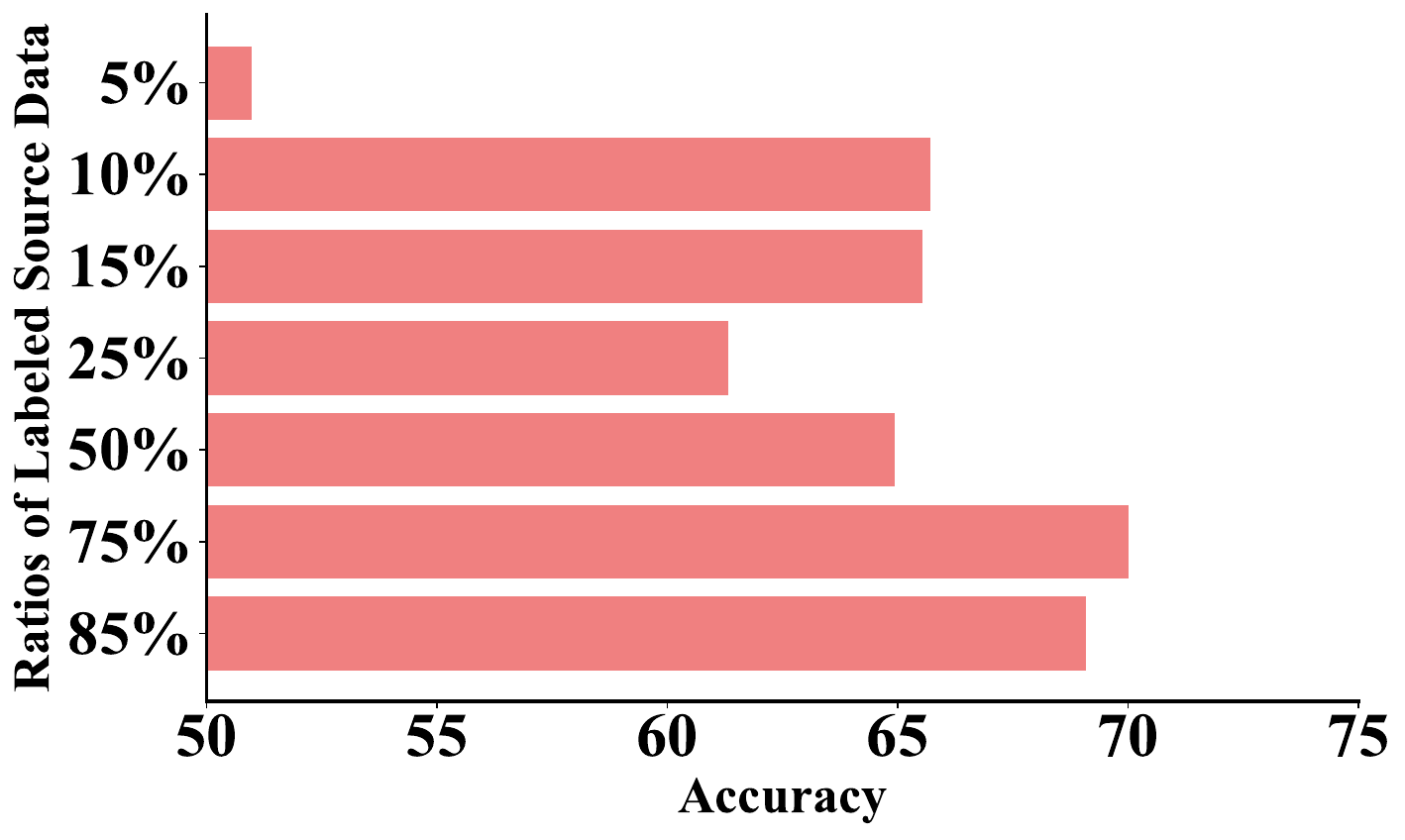}
\end{center}
\caption{The mean accuracy rates (\%) of the model with different proportions of labeled source domain data.}
\label{fig:histogram}
\end{figure}

The availability of labeled source data during semi-supervised learning is crucial in determining the effectiveness of model training, especially when working with complex EEG signals. Understanding the impact of different ratios of labeled source data helps evaluate the trade-off between data quantity and model performance, providing insights into how much labeled data is required to achieve optimal outcomes. As shown in Fig. \ref{fig:histogram}, we adjust the ratios of labeled source data $\tau\%$ from 5\% to 85\%.

The experimental results show that with only 5\% of the labeled source domain data, the model struggles to learn effectively, achieving a relatively low accuracy rate of 50.97\%. When the ratio increases to 10\%, there is a significant improvement in accuracy, with the model reaching an average accuracy of 65.71\%. This suggests that the model requires a minimum ratio of labeled source data to begin capturing meaningful patterns in the EEG signals. Beyond this 10\% ratio, the accuracy continues to improve steadily, reaching a plateau when 75\% is reached, with an accuracy of 70.00\%. This suggests that the model, leveraging its semi-supervised learning framework, effectively utilizes both labeled and unlabeled data to maintain steady and robust performance, even when labeled source data is relatively limited. However, when the labeled data ratio reaches 85\%, a slight decrease in accuracy is observed. This is likely due to increased variability in the labeled data, introducing inconsistencies such as annotation noise or domain shifts. As a result, the model becomes more sensitive to these variations and leads to a decline in generalization ability. This emphasizes the importance of balancing labeled and unlabeled data in semi-supervised learning.

Ten-fold cross-validation across subjects demonstrates the generalization ability of NSSI-Net. On the other hand, identifying the minimum number of training subjects remains important for computational efficiency. As shown in Fig. \ref{fig:histogram}, performance improves significantly with around 10\% labeled data (about 5 subjects) and continues to rise until about 75\% (around 40 subjects). Incorporating 10\%–25\% of subjects (5~13 in this study) into the labeled training set offers an optimal balance between accuracy and computational cost.

\subsection{The Effect of Sampling Process}
\begin{figure}
\begin{center}
\includegraphics[width=0.5\textwidth]{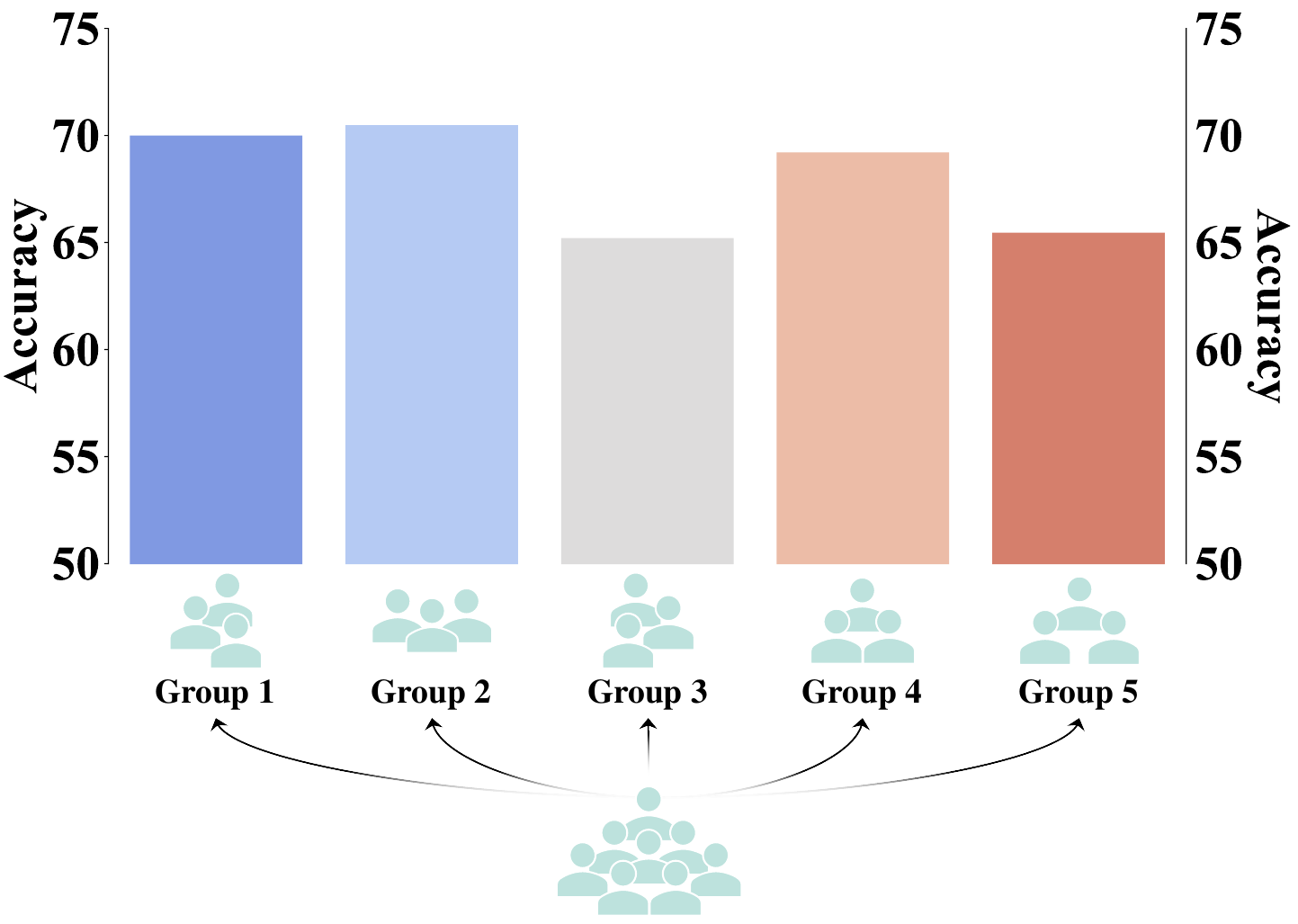}
\end{center}
\caption{The accuracy distribution across five rounds of random sampling using semi-supervised cross-subject ten-fold cross-validation.}
\label{fig:strip}
\end{figure}

To balance the data distribution and gender representation, a data sampling strategy is adopted as described in Section \ref{sec:inputData}. Here, we further evaluate the effect of this sampling process on model performance. As illustrated in Fig. \ref{fig:strip}, we randomly form different sample groups and re-implement the semi-supervised cross-subject ten-fold cross-validation for each group separately. The model performance across these sampled groups ranges from 65.21\% to 70.48\%, with corresponding standard deviations between 13.18\% and 14.59\%. The results demonstrate that the model consistently delivers stable performance across various sampling configurations, indicating its robustness to variations in data selection during the sampling process.

\subsection{The Effect of Parallel Processing in Feature Extraction}
To compare the effectiveness of sequential and parallel feature extraction, we adjust the feature extraction pipeline. For clarity, we refer to the model where spatial features are extracted first, followed by temporal features, as the \textbf{sequential model} (the original architecture). Meanwhile, we refer to the model where spatial and temporal features are extracted simultaneously as the \textbf{parallel model}. Unlike the sequential model, which maintains a symmetric CNN-BiGRU encoder and BiGRU-CNN decoder, the parallel model lacks this symmetry, requiring an alternative decoding approach with a multi-layer convolutional decoder instead. To assess model performance, we conduct cross-validation under both processing strategies. The results indicate a significant performance decline from 70.00$\pm$13.90 (sequential model) to 58.03$\pm$5.59 (parallel model). This degradation may be primarily due to compromised temporal consistency. In the sequential model, BiGRU receives structured spatial-temporal features, which facilitates effective sequential modeling. In contrast, the parallel model forces BiGRU to process raw EEG data directly. This reduces its ability to capture long-term dependencies. These findings suggest that the original sequential structure remains superior for EEG-based NSSI detection, as it better preserves hierarchical feature integration and temporal dependencies.

\subsection{Topographic Analysis of Channel Contributions to NSSI Classification}
To deepen our understanding of the neural mechanisms underlying NSSI detection, we analyze the contribution of each EEG channel to the model's ability to distinguish between DN+ and DN-. Instead of using signals from all brain regions as input, we examine the role of individual EEG channels by feeding data from each channel independently into the model. For each channel, we determine the model's classification accuracy and subsequently normalize these accuracy values across all channels.

\begin{figure*}
\begin{center}
\includegraphics[width=0.8\textwidth]{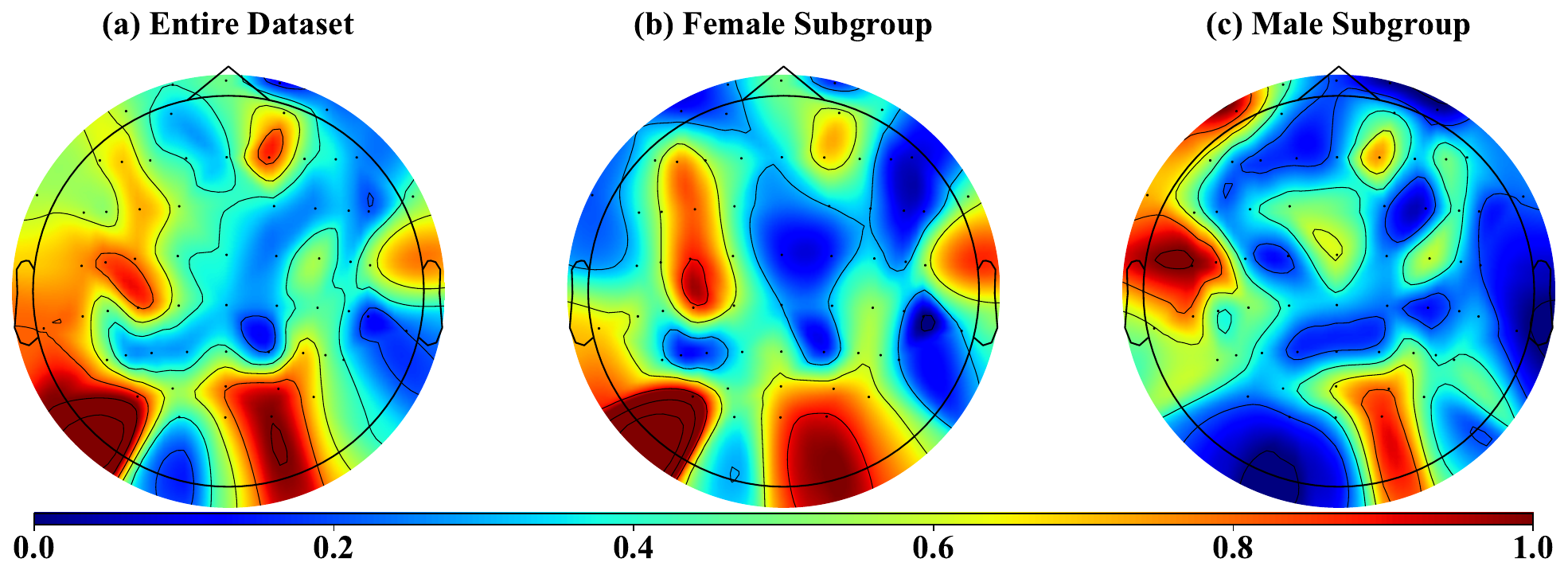}
\end{center}
\caption{The topographic maps show the contribution of each EEG channel to the model’s performance across three datasets: (a) Entire Dataset, (b) Female Subgroup, and (c) Male Subgroup. The color bar indicates the importance of each channel in distinguishing between the DN+ and DN- groups. Red represents higher importance, while blue indicates lower contribution.}
\label{fig:topo}
\end{figure*}

The normalized accuracy values are visualized in a topographic analysis, which allows us to discern spatial patterns of channel importance. As shown in the Fig. \ref{fig:topo} (a), the parietal, frontal and occipital regions demonstrate significantly higher importance in predicting NSSI behaviors compared to other brain areas. These regions are known to play key roles in emotional processing \cite{esslen2004brain, li2013abnormal, fusar2009functional}, visuospatial functions \cite{kravitz2011new,chai2024feasibility}, and self-referential thoughts \cite{northoff2006self, benoit2010think, wagner2015neural}, which could be critical in the development and maintenance of self-injurious behaviors.

When analyzing the gender-specific results, we observe the differences between the female and male subgroups. In female subgroup (Fig. \ref{fig:topo} (b)), emotional processing engages both hemispheres, with bilateral activation in key regions, suggesting a more integrated neural processing approach. In contrast, the male subgroup rely more on unilateral processing (Fig. \ref{fig:topo} (c)), with emotional responses exhibiting greater lateralized activity. These findings indicate gender-specific differences in emotional processing, with females utilizing both hemispheres and males showing more lateralized brain activity \cite{ingalhalikar2014sex}.

Gender-specific analyses are conducted to explore potential neurophysiological factors affecting model performance, with the aim of enhancing scientific understanding rather than implying normative differences. Existing literature highlights biological and cognitive variations between males and females, particularly in emotional regulation and self-referential cognition\cite{xu2023functional,sun2025gender}, which are key domains in the study of NSSI-related neural dynamics.


\subsection{Model Stability and Convergence Performance}
\begin{figure*}
\begin{center}
\includegraphics[width=0.9\textwidth]{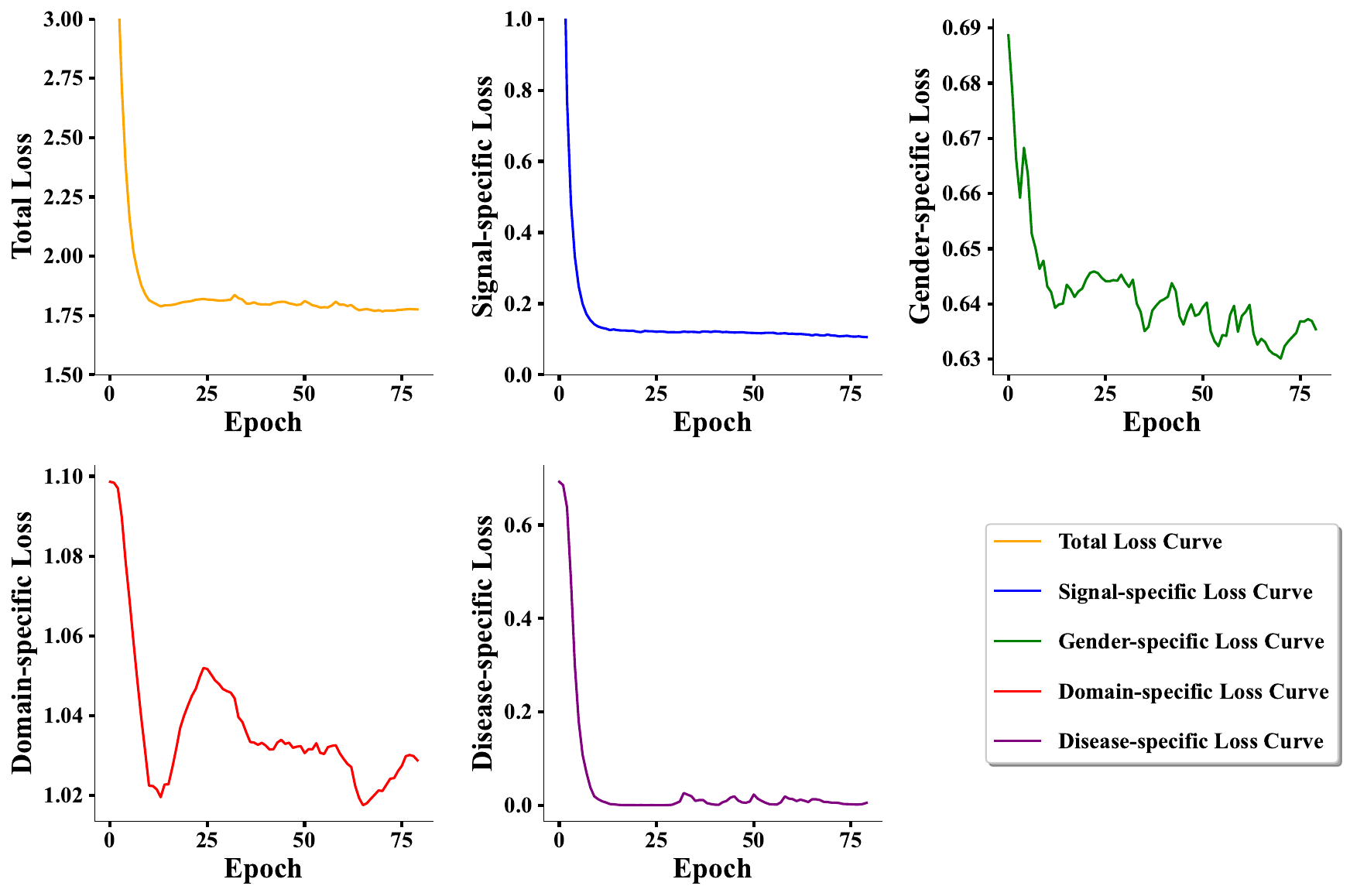}
\end{center}
\caption{The total loss (yellow) represents the sum of the individual losses, and the curves show the loss values for the signal-specific (blue), gender-specific (green), domain-specific (red), and disease-specific (purple) discriminators over 80 epochs.}
\label{fig:loss_curve}
\end{figure*}

To assess stability and convergence, we evaluate the overall performance of the designed multi-concept discriminator and the individual performance of each discriminator by examining their respective loss curves throughout the training process. We compute the average loss curves across all cross-validation folds to ensure a comprehensive evaluation of training dynamics, with the results presented in Fig. \ref{fig:loss_curve}. For the overall performance of the multi-concept discriminator, the total loss curve (yellow) exhibits smooth and stable convergence after approximately 60 epochs. It indicates that the model successfully reaches an equilibrium where further training provides minimal gains and ensures strong generalization performance. For individual discriminators, the signal-specific discriminator’s loss curve (blue) stabilizes rapidly, converging around the 15th epoch. It suggests its effectiveness in distinguishing between true and false signals early in training. The gender- (green) and domain-specific (red) discriminators exhibit mild oscillations in the initial training phase, a common characteristic of adversarial networks. Nevertheless, both show a steady decline in loss after approximately 60 epochs, indicating effective feature learning. Similarly, the disease-specific discriminator (purple) stabilizes early, similar to the signal-specific discriminator, highlighting its strong feature extraction capability.

Furthermore, we conduct comprehensive experiments with various weight configurations for the hyperparameters ($\alpha$, $\beta$, $\delta$, $\theta$) used in Eq. \ref{Eq:l_total}. Specifically, we evaluate these hyperparameters ($\alpha$, $\beta$, $\delta$, $\theta$) with the following weight ratios: 1:1:1:1, 1:1:1:2, 1:1:2:1, 1:2:1:1, and 2:1:1:1. The corresponding results are 70.00$\pm$13.90, 65.12$\pm$09.07, 69.10$\pm$10.18, 69.11$\pm$13.93, and 63.56$\pm$08.73, respectively. These experiments reveal that the model achieves optimal performance with a balanced configuration ($\alpha$:$\beta$:$\delta$:$\theta$ = 1:1:1:1). It is noted that, increasing the weights of the disease-specific and signal-specific components leads to a significant decrease in performance. This suggests that the disease-specific and signal-specific discriminators converge more quickly during training. When these components are given excessive weight, they would dominate the learning process, causing the model to focus too narrowly on certain aspects. This analysis highlights the importance of balanced weight configurations for achieving optimal performance across all discriminators.

\subsection{Conclusion}
In this paper, we introduce a novel semi-supervised framework with a multi-concept discriminator (NSSI-Net), designed for the detection of NSSI using EEG signals. NSSI-Net leverages spatial-temporal EEG patterns through a hybrid CNN-BiGRU encoder-decoder structure, which allows us to characterize the complexity of neural activities in adolescents with depression. The multi-concept discriminator is proposed to address the heterogeneous and complex nature of EEG data by focusing on multiple discriminative perspectives. It specifically focuses on capturing signal-specific, gender-specific, domain-specific, and disease-specific characteristics, enhancing the model’s ability to accurately classify NSSI-related features. Based on a self-collected dataset of 114 adolescents diagnosed with depression, the performance of the proposed NSSI-Net is validated using a semi-supervised cross-subject cross-validation approach. Comprehensive analysis of model components and hyperparameters are carefully conducted to assess the framework’s robustness. The results show that NSSI-Net achieves high efficiency and accuracy in detecting NSSI and provides interpretable insights into the neural mechanisms associated with NSSI. 
Future work would focus on enhancing the model’s ability to handle class imbalances and sample size limitations, including increasing the male sample size and expanding the dataset to ensure consistent performance across a broader range of real-world applications.


\section*{References}
\bibliographystyle{IEEEtran}
\bibliography{references}

\end{document}